\documentclass[]{spie}  

\usepackage{amsmath,amsfonts,amssymb}
\usepackage{graphicx}
\usepackage[colorlinks=true, allcolors=blue]{hyperref}

\usepackage{graphicx}
\graphicspath{{./images/}}
\usepackage{color}
\usepackage{xcolor, soul}
\sethlcolor{red}
\usepackage{amsmath}
\usepackage{dsfont}
\usepackage{hyperref}
\usepackage[capitalise]{cleveref}
\usepackage{subcaption}
\usepackage{placeins}
\usepackage{footnote}
\usepackage{listings}
\usepackage[T1]{fontenc}
\PassOptionsToPackage{hyphens}{url}
\usepackage{url}
\usepackage{flushend}
\definecolor{codegreen}{rgb}{0,0.6,0}
\definecolor{codegrey}{rgb}{0.5,0.5,0.5}
\definecolor{backcolour}{rgb}{1,1,1}
\definecolor{orange}{rgb}{1,0.5,0}
\usepackage[printonlyused]{acronym}
\usepackage{multirow}
\usepackage{tabularx}
\newcolumntype{C}[1]{>{\centering\arraybackslash}p{#1}}
\newcolumntype{L}[1]{>{\arraybackslash}p{#1}}
\usepackage{diagbox}

\usepackage{booktabs}
\newcommand*{\addheight}[2][0.5ex]{%
  \raisebox{0pt}[\dimexpr\height+(#1)\relax]{#2}%
}

\title{Advanced Deep Learning Architectures for Accurate Detection of Subsurface Tile Drainage Pipes from Remote Sensing Images}

\author[a]{Tom-Lukas~Breitkopf$^*$}
\author[a]{Leonard~Hackel$^*$}
\author[a]{Mahdyar~Ravanbakhsh}
\author[b]{Anne-Karin~Cooke}
\author[b]{Sandra~Willkommen}
\author[b]{Stefan~Broda}
\author[a]{Begüm~Demir}
\affil[a]{Technische Universität Berlin, 10623 Berlin, Berlin, Germany}
\affil[b]{Bundesanstalt für Geowissenschaften und Rohstoffe, 13593 Berlin, Berlin, Germany}

\authorinfo{Further author information: (Send correspondence to Leonard Hackel)\\Leonard Hackel: \href{mailto://l.hackel@tu-berlin.de}{E-mail: l.hackel@tu-berlin.de}}

\begin{document} 
\maketitle
\def\thefootnote{*}\footnotetext{These authors contributed equally to this work}

\acrodef{DL}{deep learning}
\acrodef{ML}{machine learning}
\acrodef{RS}{remote sensing}
\acrodef{VT}{Visual Transformer}
\acrodef{ReLU}{rectified linear unit}
\acrodef{ASPP}{atrous spatial pyramid pooling}
\acrodef{AG}{attention gate}
\acrodef{BGR}[BGR]{Deutsche Bundesanstalt für Geowissenschaften und Rohstoffe}
\acrodef{GPR}{ground-penetrating radar}
\acrodef{FCN}{fully convolutional network}
\acrodef{IoU}{intersection over union}
\acrodef{TP}{true positive}
\acrodef{FP}{false positive}
\acrodef{TN}{true negative}
\acrodef{FN}{false negative}
\acrodef{NLP}{natural language processing}
\acrodef{UAV}{unmanned areal vehicle}
\acrodef{AUC}{Area under ROC Curve}

\begin{abstract}
Subsurface tile drainage pipes provide agronomic, economic and environmental benefits. By lowering the water table of wet soils, they improve the aeration of plant roots and ultimately increase the productivity of farmland. They do however also provide an entryway of agrochemicals into subsurface water bodies and increase nutrition loss in soils. For maintenance and infrastructural development, accurate maps of tile drainage pipe locations and drained agricultural land are needed. However, these maps are often outdated or not present. Different \ac{RS} image processing techniques have been applied over the years with varying degrees of success to overcome these restrictions.
Recent developments in deep learning (DL) techniques improve upon the conventional techniques with machine learning segmentation models. In this study, we introduce two DL-based models: i) improved U-Net architecture; and ii) Visual Transformer-based encoder-decoder in the framework of tile drainage pipe detection. Experimental results confirm the effectiveness of both models in terms of detection accuracy when compared to a basic U-Net architecture. Our code and models are publicly available at \url{https://git.tu-berlin.de/rsim/drainage-pipes-detection}.
\end{abstract}


\section{INTRODUCTION}
\label{sec:intro}  
Subsurface tile drainage pipes are an essential agricultural element, maintaining or improving water management and crop yields of farm land. Depending on location and respective hydro-climatological conditions, significant parts of farm land are equipped with tile drainage systems (e.g. in Denmark $\sim50\%$ of the agricultural area [\citenum{Motarjemi2021drainage_mla}]). Multiple installation patterns of subsurface drainage system are in use like herringbone, double main, parallel, targeted and complex patterns (see [\citenum{koganti2020mapping}] for more information).

\begin{figure}
    \centering
    \includegraphics[scale=0.45]{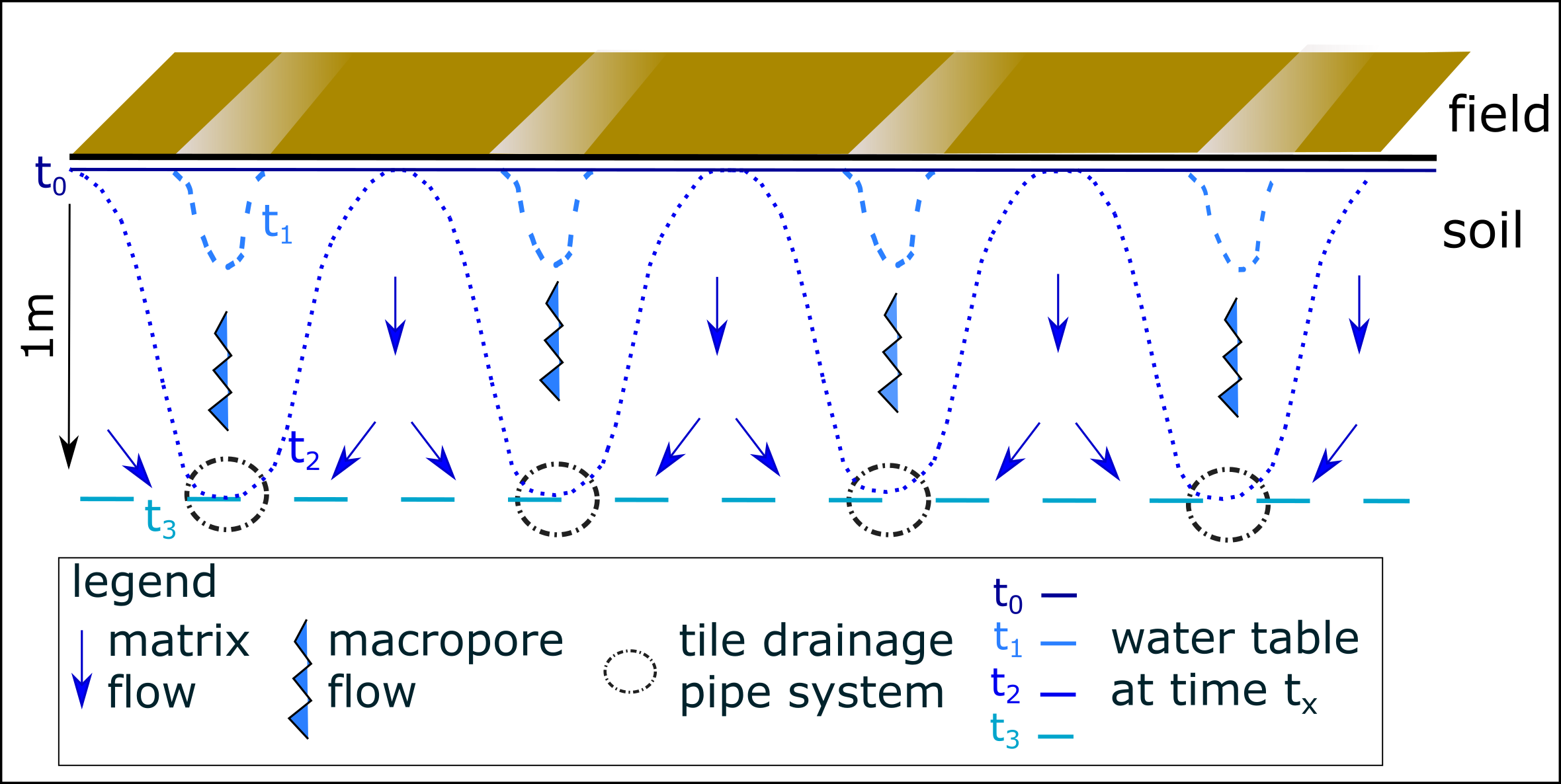}
    \caption{Pictorial representation of drainage pipes and the impact on soil color on the surface [\citenum{song2021detecting}].}
    \label{fig:drainage_pipes2}
\end{figure}

Besides their positive effects, subsurface drainage pipes also boost -- in high pulse-like signals -- nutrient and pesticide loss of soils with short retention time to surface waters. Especially in areas where surface runoff cannot reach receiving waters and infiltrates into depressions, macropore transport to tile drainage pipes was shown as a dominant pesticide loss pathway of agrochemicals into surface water bodies [\citenum{Leu2004herbicideloss}], posing risk to both human and ecosystem health. In order to improve the understanding of agro-chemical dynamics in soils, to calculate tile drainage catchment areas and loads for wetland dimension planning, as well as to prepare reliable eco-hydrological models, it is thus crucial to identify the location of drainage pipes [\citenum{DeSchepper2017drainag_model}]. One typical example for parallel tile drainage pipes in flat lowland regions is depicted in \cref{fig:drainage_pipes2}. The pipes are roughly ten centimeters in diameter and installed at about one meter below surface [\citenum{song2021detecting}]. Precipitation (at $t_0$) infiltrates rapidly through the soil above the drainage pipes ($t_1$ and $t_2$) due to preferential flow through especially macropores, leading to relatively low (close to drainage pipes) and relatively high (farther away from the pipes) soil moisture. As the surface albedo increases with decreasing soil moisture, those differences in soil moisture become visible. As shown in \cref{fig:drainage_pipes2}, the soil above the drainage pipes appears brighter after a precipitation event, thus producing distinct features visible at the surface [\citenum{koganti2020mapping, song2021detecting}].

The lack of documentation of installed drainage pipes has motivated researchers to explore other detection methods, such as thermal images and ground-penetrating radar (GPR) [\citenum{Allred2020thermal}]. Their practical applicability is however limited by the restricted data availability. Conventional image processing methods (e.g. edge detection, decision tree classification and image differencing [\citenum{gokkaya2017subsurface}]) have been applied to remote sensing images of bare soils, obtained after precipitation events [\citenum{Kratt2020FieldTT}]. Machine learning methods in particular DL-based methods have recently shown good performance on drainage pipe detection from remote sensing images [\citenum{Cho_2019_ML_drainage, OHara2020_ML_drainage, song2021detecting}]. 
As an example, in [\citenum{song2021detecting}] a DL-based method is introduced using the deep autoencoder-decoder U-Net architecture [\citenum{ronneberger2015u}] to detect drainage pipe from remote sensing images. The U-Net architecture outperforms other remote sensing image-based approaches using edge detection. The authors suggest that even though the model shows good performance, it may still be improved by tuning hyperparameters and using a more diverse training set [\citenum{song2021detecting}].

In this work we aim to improve on the results of the model introduced in [\citenum{song2021detecting}] and introduce two existing DL-based models to the problem of drainage pipe detection: i) a modified U-Net architecture with multiple refinements (denoted as improved U-Net); and ii) a visual transformer-based encoder-decoder architecture with skip connections (denoted as TransUNet).

\section{METHOD}\label{sec:unet}

We adapt two DL-based models to the task of semantic segmentation for tile drainage pipe detection, one based on an improved U-Net and one based on a Visual Transformer. Both models take as input an image $x$ with $C$ channels and a $H\times W$ resolution (height and width) and output a grayscale image of the same resolution. The gray value of a pixel in the output represents the probability of a drainage pipe being in the location of that pixel. Formally, the goal is to predict a pixel-wise mapping $M$ with $x \mapsto M(x)$ where $x \in \mathds{R}^{H\times W\times C}$ and $M(x) \in \mathds{R}^{H\times W\times 1}$. 
The mapping function is trained by minimizing the dice loss function $L_{Dice}$ as defined in [\citenum{song2021detecting}]. The details of each method are presented in the following subsections.


\begin{figure}
    \centering
    \includegraphics[width=.8\columnwidth]{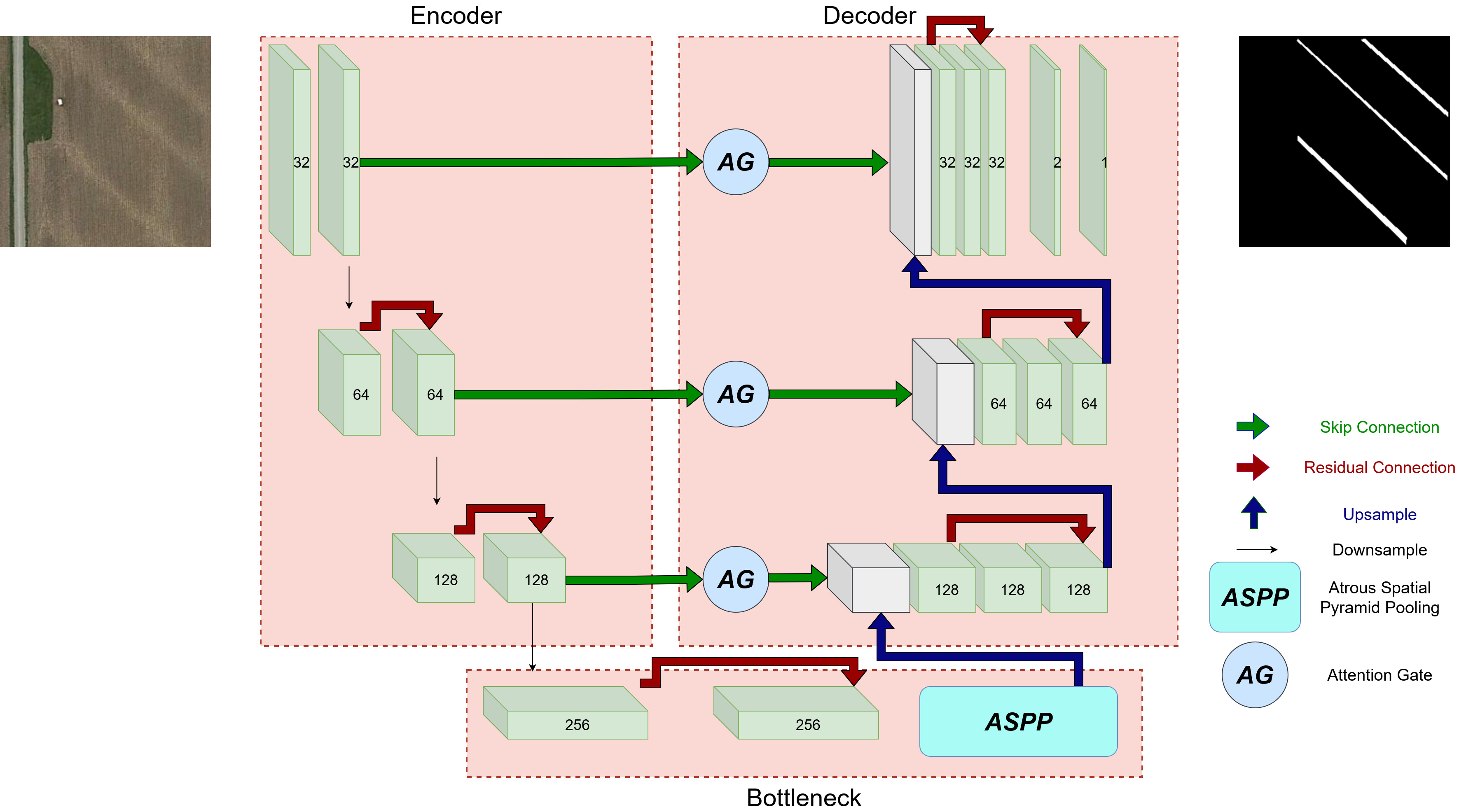}
    \caption{Architectures of the introduced improved U-Net is based on a four-layer U-Net consisting of an encoder and a decoder part and is enhanced with residual blocks, \ac{AG} and \ac{ASPP} modules. The 'bottleneck' is the most contracted part with the highest number of convolutional kernels [\citenum{augustauskas2020improved}].}
    \label{fig:improved_unet_architecture}
\end{figure}
\begin{figure}
    \centering
    \includegraphics[width=.8\columnwidth]{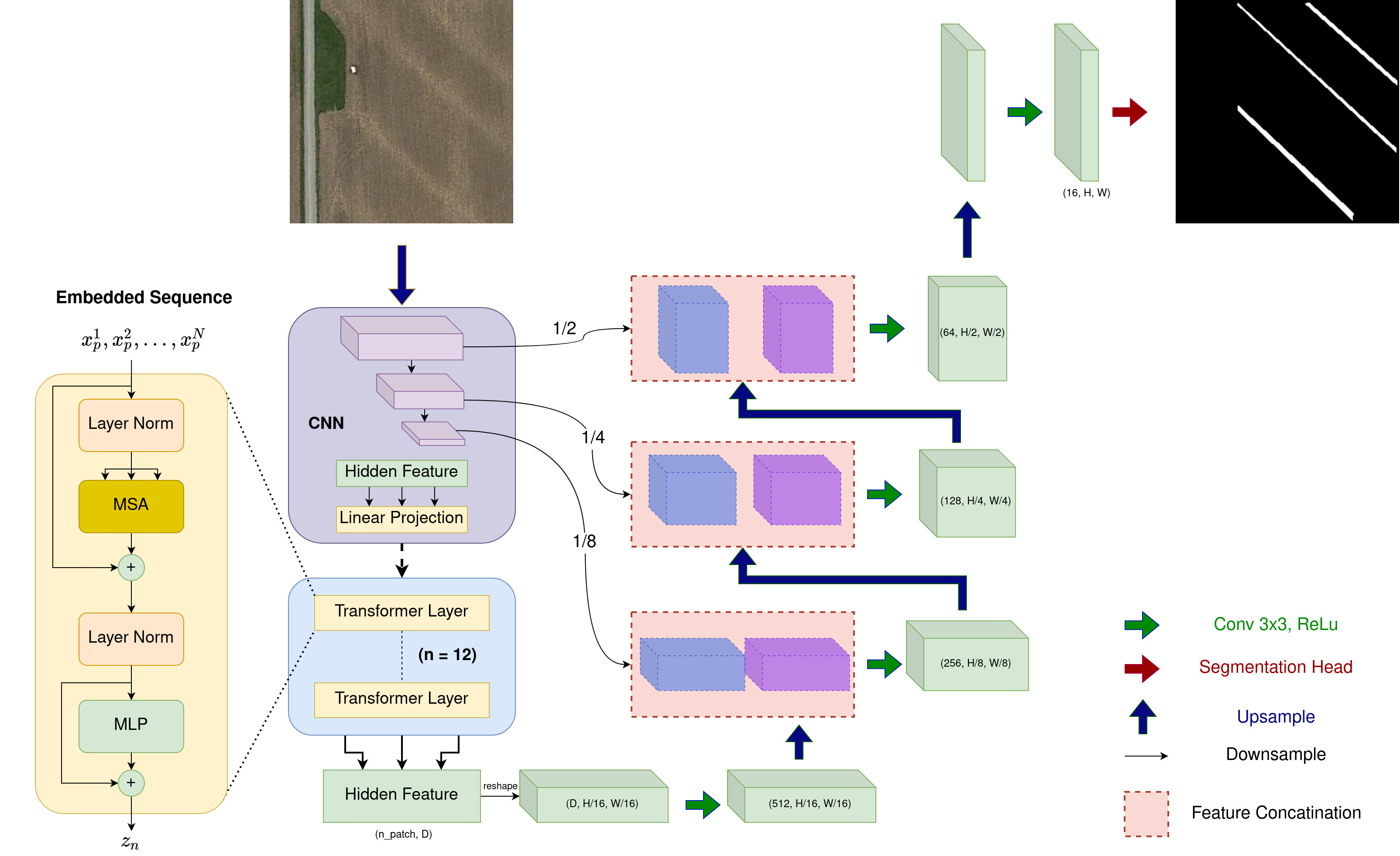}
    \caption{Architectures of the introduced TransUNet network: A CNN is used for low level feature extraction, followed by a transformer encoder for high level features. To improve the location accuracy of the upscaling decoder, skip connections are added from the CNN to the decoder layers [\citenum{chen2021transunet}].}
    \label{fig:transunet_arch}
\end{figure}


\subsection{Improved U-Net}\label{ssec:unet}
The U-Net architecture [\citenum{ronneberger2015u}] is a fully convolutional auto-encoder-decoder (FCN) networks, which contains a contracting and an expanding path. The contracting path captures context, whereas the expanding path allows precise localization. Both paths are in a way symmetric to each other. 
The U-Net can make use of skip connections that makes it possible for the model to be trained with few images. Detecting tile drainage pipes from RGB remote sensing images using a deep U-Net architecture was proposed in [\citenum{song2021detecting}]. Since the U-Net architecture used in [\citenum{song2021detecting}] is a basic U-Net, it is not able to utilize multiscale contexts in the latent space at the bottleneck or emphasize relevant and deemphasize irrelevant information in the residual connections. These issues get addressed in the improved U-Net architecture. In [\citenum{augustauskas2020improved}] several U-Net architectures with several adaptions are introduced, improving the image segmentation performance without significant computational overhead. In this work we investigate one of these architectures and adapt it to the task of drainage pipe detection. This improved U-Net architecture incorporates several additional modules w.r.t. the basic U-Net. As it is shown in \cref{fig:improved_unet_architecture} the core of the architecture is a four-layer U-Net network using 3$\times$3 kernels and a stride of one for all convolutions. Padding was added, so that input and output dimensions match and batch normalization was added between each convolutional layer and its activation function (except for the last output layer), to make the network more stable. 
The main architectural changes with regard to the basic U-Net architecture are: i) added residual connections; ii) atrous spatial pyramid pooling (ASPP); and iii) attention gate (AG) modules [\citenum{augustauskas2020improved}].
Residual connections addresses the degradation problem, which occurs when the performance of a network degrades with increasing depth, by introducing shortcut connections, skipping one ore more layers. Residual blocks thus allow to gain accuracy through increased depth, while being easy to optimize [\citenum{he2016deep}].
In the \ac{ASPP} module a sequence of convolutions with different dilation rates is performed [\citenum{augustauskas2020improved}]. The dilated convolutions allow to expand the receptive field without losing resolution. A module using dilated convolutions is thus able to extract multi-scale information [\citenum{yu2016multiscale}].
The \ac{AG} modules learn to focus on the target structure without the need of further supervision [\citenum{schlemper2019attention}]. 
In this way they highlight important features, while suppressing irrelevant ones [\citenum{augustauskas2020improved}].
More detailed information on residual blocks, the \ac{ASPP} and \ac{AG} modules and the way they are combined together in the improved U-Net model (ResU-Net + ASPP + AG) can be found in [\citenum{he2016deep, augustauskas2020improved, yu2016multiscale, schlemper2019attention}].

\subsection{TransUNet} \label{ssec:vt}
The TransUNet utilizes a visual transformer-based architecture [\citenum{dosovitskiy2020image}] and divides a given image into a sequence of tokens and processes it via a stack of combinations of a multi-head attention layer similar to what was described in \cref{ssec:unet} and a fully connected layer. When dividing the image into small patches the image $x$ can be described as a sequence of $N$ non-overlapping patches $\{x_p^i \in \mathds{R}^{P^2\cdot C} | i = 1, \dots, N\}$ with patch size $P \times P$ and $N = \frac{HW}{P^2}$ patches. The number of patches is equal to the length of the input sequence. 
Parts of the patches are weighted using multiple attention heads, where the patches are treated like in a sequence problem. The fully connected layer combines these weighted results in combination with residual connections to produce higher level features. 
The encoder architecture is refined with skip connections to retain low-level features for tasks like segmentation as proposed in [\citenum{chen2021transunet}]. This results in a U-Net-like architecture. 
The skip connections are inserted between the patch encoders and the upscaling layers. Additionally, in this paper we use a simple convolution neural network (CNN) in the patch encoding module as shown in \cref{fig:transunet_arch}. 
The feature maps of the convolutions are the input of the encoder and are directly connected (via skip connections) to the upscaling decoder of high-level feature maps created by the encoder. 
This results in the decoder being able to combine high-level features form the transformer encoder and low-level features from the convolutions.

As it is shown in [\citenum{vaswani2017attention}], the visual transformers can be faster than convolutional networks in the inference time, however they still take very long to train. Therefore using pretrained weights and only fine-tuning the networks is the approach taken in this work for the TransUNet model.

\section{EXPERIMENTAL RESULTS}\label{sec:experimental_results}
Experimental analyses were conducted on a dataset of RS images proposed in [\citenum{song2021detecting}]. The dataset contains 513 RGB images with pixel-level labels (i.e. drained, not drained) acquired from Sunbury, Ohio, USA. The images are in the size of 256$\times$256 pixels with a spatial resolution of 30 cm and three channels (RGB). The images were obtained using Google Earth, and the ground truth images were generated through manual annotation. The dataset was divided into 256 training images, which are augmented to a total of 3072, and 257 validation images. The augmentation includes: horizontal and vertical flips and random rotation, brightness adjustment and zooming [\citenum{song2021detecting}].

In order to evaluate the improved U-Net and TransUNet architectures we compare them with the model used in [\citenum{song2021detecting}]. This baseline model consists of a basic U-Net architecture with only skip connections added to allow residual learning, containing around 0.5 million parameters. More detailed information can be found in [\citenum{song2021detecting}]. 
As no pretrained model was provided for [\citenum{song2021detecting}], we trained the basic U-Net using the original code, with only a slight adaption to the computation of $L_{Dice}$ to make the training more stable and allow for fair comparison with the other models. The basic U-Net was trained for 500 epochs with early stopping on the validation loss and 50 epochs patience, until it reached a training loss of 0.37 and a validation loss of 0.46 (in [\citenum{song2021detecting}] a training loss of 0.28 and a validation loss of 0.42 was achieved).
The improved U-Net was trained for 100 epochs. The initial learning rate set to 0.001, which reduced by half every 16 epochs. The model contains roughly one million parameters.
The TransUNet was trained for a total of 150 epochs. The learning rate was calculated as $lr = 0.01 \cdot (1 - \frac{i}{E})^{0.9}$, where $E$ notates the maximum epoch 150 and $i$ notates the current epoch. The Transformer backbone was pretrained with an ResNet-50 ViT-B/16 hybrid [\citenum{chen2021transunet}] model on ImageNet [\citenum{deng2009imagenet}]. As a loss function a combination of 50\% $L_{Dice}$ and 50\% cross entropy loss was used. The full model contains 105 million parameters.

To compare the performance of the different models, the \ac{IoU} and the dice coefficient [\citenum{dice1945measures}] are used. For both the \ac{IoU} and the dice coefficient a threshold is needed to be applied before the computation. We followed the evaluation pipeline introduced in [\citenum{song2021detecting}], where the results are computed by applying all the thresholds in the range from 0.1 to 0.95 in steps of 0.05 and averaging the results.
The dice coefficient and the \ac{IoU} however are sensitive to even small displacements in the prediction, as they compute a pixel-level scoring. 
To evaluate the performance of the models in operational use cases, where not the exact location of drainage pipes is of interest, but only the information if a certain area is drained or not, we introduce new measures. For that the 256$\times$256 binary ground truth and output images are transformed to 3$\times$3 images. 
The resulting 9 patches are classified into none, low, middle and high, indicating the amount of drainage in the patch, by applying thresholds on the amount of positive pixels within a patch. The thresholds are set, based on the deviation of the amount of positive pixels in all ground truth images containing any positive pixels. The threshold for a pixel to be interpreted as positive is set to 0.5.
A transformation of a high resolution ground truth image to a low resolution image can be seen in \cref{fig:high_low_conv}. By a patch-wise comparison of the resulting ground truth and output images, a confusion matrix can be computed and based on that precision and recall values can be determined for the four different classes.

\def \predictionTableImageWidth {0.16} 
\def \evaluationImageWidth {0.64} 

\begin{figure}
\centering
\begin{subfigure}{0.33\columnwidth}
  \centering
  \frame{\includegraphics[width=\evaluationImageWidth\linewidth]{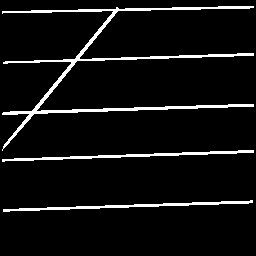}}  
  \caption{HR ground truth (256$\times$256, 8bit grayscale)}
  \label{sfig:hr_gt}
\end{subfigure}
\hspace{15pt}
\begin{subfigure}{0.33\columnwidth}
  \centering
  \frame{\includegraphics[width=\evaluationImageWidth\linewidth]{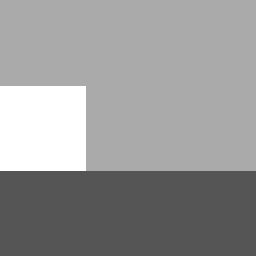}}  
  \caption{LR ground truth (3$\times$3, 2bit grayscale)}
  \label{sfig:lr_gt}
\end{subfigure}
\caption{Conversion from high resolution (HR) ground truth to low resolution (LR) ground truth.}
\label{fig:high_low_conv}
\end{figure}

\newcommand{\tablemode}{booktabs} 

\begin{table}[b]
    \centering
    \caption{$L_{Dice}$, dice coefficient and \ac{IoU} of the basic U-Net, improved U-Net and TransUNet on the validation set of the data set provided in [\citenum{song2021detecting}].}
    \label{tab:dice_iou}
    \normalsize
    \begin{tabular}{lrrr}
        \textbf{Model} & \textbf{$L_{Dice}$} &\textbf{Dice coefficient} & \textbf{\ac{IoU}} \\
        \toprule
        Basic U-Net [\citenum{song2021detecting}] & \textbf{0.46} & 0.55 & 0.76 \\
        Improved U-Net & 0.79 & 0.68 & 0.80 \\
        TransUNet & 0.79 & \textbf{0.70} & \textbf{0.81} \\
        \bottomrule
    \end{tabular}
\end{table}
\begin{table}
    \centering
    \normalsize
    \caption{Precision and recall obtained by basic U-Net, improved U-Net and TransUNet associated with the different classes.}
    \label{tab:pr_comp}
    \begin{tabular}{llrr}
         \textbf{Model} & \textbf{Class} & \textbf{Precision} & \textbf{Recall}\\
         \toprule
         \multirow{4}{*}{Basic U-Net [\citenum{song2021detecting}]} & none & 0.96 & 0.88 \\
         & low & 0.46 & 0.49 \\
         & middle & 0.49 & 0.65 \\
         & high & 0.17 & \textbf{0.32} \\
         \cmidrule(lr){1-4}
         \multirow{4}{*}{Improved U-Net}& none & \textbf{0.99} & 0.93 \\
         & low & 0.62 & \textbf{0.81} \\
         & middle & \textbf{0.71} & 0.68 \\
         & high & 0.33 & 0.13 \\
         \cmidrule(lr){1-4}
         \multirow{4}{*}{TransUNet}& none & \textbf{0.99} & \textbf{0.94} \\
         & low & \textbf{0.64} & 0.77 \\
         & middle & 0.68 & \textbf{0.72} \\
         & high & \textbf{0.69} & 0.29 \\
         \bottomrule
    \end{tabular}
\end{table}

\begin{figure*}[b]
    \centering
    \begin{tabular}{ccccc}
      \addheight{\includegraphics[width=\predictionTableImageWidth\linewidth]{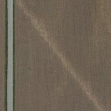}} &
      \addheight{\includegraphics[width=\predictionTableImageWidth\linewidth]{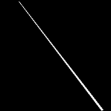}} &
      \addheight{\includegraphics[width=\predictionTableImageWidth\linewidth]{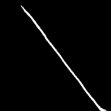}} &
      \addheight{\includegraphics[width=\predictionTableImageWidth\linewidth]{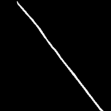}} & 
      \addheight{\includegraphics[width=\predictionTableImageWidth\linewidth]{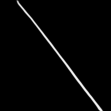}} \\
      \addheight{\includegraphics[width=\predictionTableImageWidth\linewidth]{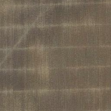}} &
      \addheight{\includegraphics[width=\predictionTableImageWidth\linewidth]{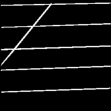}} &
      \addheight{\includegraphics[width=\predictionTableImageWidth\linewidth]{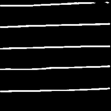}} &
      \addheight{\includegraphics[width=\predictionTableImageWidth\linewidth]{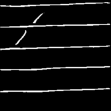}} & 
      \addheight{\includegraphics[width=\predictionTableImageWidth\linewidth]{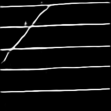}} \\
      \addheight{\includegraphics[width=\predictionTableImageWidth\linewidth]{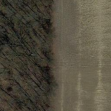}} &
      \addheight{\includegraphics[width=\predictionTableImageWidth\linewidth]{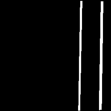}} &
      \addheight{\includegraphics[width=\predictionTableImageWidth\linewidth]{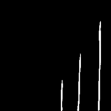}} &
      \addheight{\includegraphics[width=\predictionTableImageWidth\linewidth]{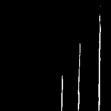}} & 
      \addheight{\includegraphics[width=\predictionTableImageWidth\linewidth]{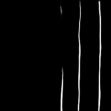}} \\
      \footnotesize (a) & (b) & (c) & (d) & (e)
    \end{tabular}
    \caption{(a) Aerial images and their (b) ground truth maps; and prediction maps obtained by using (c) basic U-Net; (d) improved U-Net; and (e) TransUNet.}
    \label{fig:prediction_matrix}
\end{figure*}

The quantitative results of the experiments can be seen in \cref{tab:dice_iou}. The basic U-Net model reaches the best validation $L_{Dice}$ loss of the three models. In terms of the dice coefficient and the \ac{IoU} score however the investigated improved U-Net and TransUNet models surpass the basic model, with the TransUNet outperforming the basic and improved U-Net for both metrics. As an example, the TransUNet obtained 5\% and 1\% higher \ac{IoU} score than the basic and improved U-Net, respectively. Qualitative results can be seen in \cref{fig:prediction_matrix}. 
The first image shows an example, where all the models create a near-perfect prediction. All the models are in that case able to distinguish between the drainage pipe and the road, even though they both appear as a straight line in the image. 
The second image shows a tendency of the improved U-Net to produce scattered lines. The drainage pipe running diagonally through the top left corner of the image is detected only partially. The TransUNet detects this drainage pipe in its full length. Instead of missing small parts of the drainage pipe it actually predicts very small features, where there should not be any. The basic U-Net fails to detect this drainage pipe completely. 
In the third image all models have some false positive predictions, detecting a drainage pipe where there should not be one. The surface feature producing the false positive is however hard to distinguish from a feature actually caused by a drainage pipe, even for the human eye. Again it can be seen, that the TransUNet detects the drainage pipe continuously in their full length compared to the basic and the improved U-Net, but also shows a little more false positives.

We also evaluate the drainage pipe detection models based on a patch-level classification. The confusion matrices of the different models and a comparison of the precision and recall can be seen in \cref{tab:pr_comp} and \ref{tab:cms}.
Both the improved U-Net and the TransUNet perform well for the none, low and middle class. For those three classes both the precision and the recall lay above 60\%. For the basic U-Net however the precision and recall in some cases are below 50\%. Only for the high class, all models perform poorly. This might be related to the low frequency with which the label appears in the data set. 
Furthermore, by analyzing the results in the confusion matrix, one can see that the predictions obtained from all three models are not too far off. As an example, patches that are highly drained are classified as middle. Similarly, patches that were wrongly classified as high are usually in the middle class.
\begin{table}
\centering
    \caption{Confusion matrices of the basic U-Net, improved U-Net architecture and TransUNet.}
    \label{tab:cms}
    \begin{tabular}{ll|rrrr}
         Model & \diagbox{GT}{pred.} &   None    &   Low     &   Middle  &   High    \\
         \toprule
         \multirow{4}{*}{Basic U-Net [\citenum{song2021detecting}]} & None   &   1433    &   185     &   19      &   0       \\
                                   & Low    &   33      &   181     &   154     &   4       \\
                                   & Middle &   22      &   26      &   180     &   45      \\
                                   & High   &   2       &   5       &   14      &   10      \\
         \cmidrule(lr){1-6}
         \multirow{4}{*}{Improved U-Net} & None   &   1530    &   105     &   2       &   0       \\
                                   & Low    &   16      &   301     &   55      &   0       \\
                                   & Middle &   6       &   74      &   185     &   8      \\
                                   & High   &   1       &   6       &   20      &   4      \\
         \cmidrule(lr){1-6}
         \multirow{4}{*}{TransUNet} & None   &   1539    &   92      &   6       &   0       \\
                                   & Low    &   15      &   288     &   69      &   0       \\
                                   & Middle &   6       &   65      &   198     &   4     \\
                                   & High   &   1       &   4       &   17      &   9       \\
        \bottomrule
    \end{tabular}
\end{table}

\section{CONCLUSION AND DISCUSSION}\label{sec:conclusion}
In this paper, we have adapted two DL-based models: i) an improved U-Net architecture; and ii) a Visual Transformer (TransUNet) to detect drainage pipes from remote sensing images. Both models take as input RGB images and provide as output a pixel-level prediction of a drainage pipe locations. We have demonstrated that the two models outperform the previous state-of-the-art method in terms of dice coefficient and IoU. In our experiments, the TransUNet has shown the best performance. However, it contains two orders of magnitude as many parameters as the improved U-Net model and thus takes much longer to train while only offering a slight improvement. The decision of which model to use is thus a tradeoff depending on the use case and the available resources. Besides the pixel-level evaluation, we have evaluated the models based on patch-wise thresholding. For real-world scenarios, the model has to predict if a certain area is drained and if so, to what extent. To this end, we divided the pixel-level predictions into four classes based on the amount of positive pixels: high, middle, low and none. For all but the 'high' class, the improved U-Net and the TransUNet outperform the basic U-Net in terms of precision and recall scores. This demonstrates the superiority of the investigated models for drainage pipe detection with respect to the basic U-Net model. 
As future work, we plan to incorporate more diverse data from different domains under different acquisition conditions, which can help the models to learn more diverse features and possibly improve the prediction quality. Furthermore, such an approach could be used to investigate whether the current restrictive soil conditions are required for image acquisition.

\acknowledgments 
This work is funded by the European Research Council (ERC) through the ERC-2017-STG BigEarth Project under Grant 759764 and by the German Ministry for Education and Research as BIFOLD - Berlin Institute for the Foundations of Learning and Data (01IS18025A).  

\bibliography{report} 
\bibliographystyle{spiebib} 

\end{document}